\documentclass[conference]{IEEEtran}
\IEEEoverridecommandlockouts
\usepackage{cite}
\usepackage{amsmath,amssymb,amsfonts}
\usepackage{algorithmic}
\usepackage{graphicx}
\usepackage{textcomp}
\usepackage{xcolor}
\usepackage{epsfig}

 \usepackage{amssymb}
\usepackage{multirow}
\usepackage{pifont}
\usepackage{graphicx}
\usepackage{float} 
\usepackage{amsmath}
\usepackage{textcomp}
\usepackage{comment}
\usepackage{verbatim}
\usepackage{xcolor}
\usepackage{colortbl}

\def\BibTeX{{\rm B\kern-.05em{\sc i\kern-.025em b}\kern-.08em
    T\kern-.1667em\lower.7ex\hbox{E}\kern-.125emX}}
\begin{document}

\title{HpEIS: Learning Hand Pose Embeddings for Multimedia Interactive Systems
\thanks{ Moodagent partially funded S.X. RM-S \& C.K. acknowledge funding by EPSRC projects EP/R018634/1, EP/MO1326X/1, EP/Y029178/1 and EP/T021020/1. Xuri Ge and Roderick Murray-Smith are corresponding authors. Email: \{s.xu.1, x.ge.2\}@research.gla.ac.uk, \{chaitanya.kaul, roderick.murray-smith\}@glasgow.ac.uk. }
}

\author{Songpei Xu, Xuri Ge, Chaitanya Kaul, Roderick Murray-Smith \\
\IEEEauthorblockA{\textit{School of Computing Science,} 
\textit{University of Glasgow}, Glasgow, UK} 
}

\maketitle

\begin{abstract}
We present a novel \textbf{H}and-\textbf{p}ose \textbf{E}mbedding \textbf{I}nteractive \textbf{S}ystem (\textbf{\textit{HpEIS}}) as a virtual sensor, which maps users' flexible hand poses to a two-dimensional visual space using a Variational Autoencoder (VAE) trained on a variety of hand poses. HpEIS enables visually interpretable and guidable support for user explorations in multimedia collections, using only a camera as an external hand pose acquisition device. 
We identify general usability issues associated with system stability and smoothing requirements through pilot experiments with expert and inexperienced users. We then design stability and smoothing improvements, including hand-pose data augmentation, an anti-jitter regularisation term added to loss function, stabilising post-processing for movement turning points and smoothing post-processing based on One Euro Filters. 
In target selection experiments (n=12), we evaluate HpEIS by measures of task completion time and the final distance to target points, with and without the gesture guidance window condition. 
Experimental responses indicate that HpEIS provides users with a learnable, flexible, stable and smooth mid-air hand movement interaction experience.
\end{abstract}


\section{Introduction}
\label{sec:intro}

    Interaction based on mid-air gestures or hand poses has a broad scope of applications as an alternative to screen touching or external physical device dependencies in cases of diverse hand conditions, e.g. wet or greasy hands, for multimedia industry applications, e.g. music and movie multimedia system interaction. 
    However, given numerous challenges and complexities of this task, there are still no robust solutions to study interactions in this domain. 
    To date, existing methods \cite{du2022mobile,tseng2023fingermapper,xu2023continuous} struggle to create naturally-interacting strategies to link real-world mid-air hand poses with virtual multimedia. Further interaction issues include jitter, instability and non-smoothness of dynamic movements not captured. 
    Among them, representing mid-air hand poses in an embedding space for interaction is a straightforward yet effective solution to this task, which serves as a heuristic for connecting multimedia.

    In the past decade, mid-air hand movement interaction and control have become a popular research topic in the human-computer interaction (HCI) community. 
    Most interaction and control methods \cite{du2022mobile,tseng2023fingermapper} used black-box processing to directly acquire corresponding feedback from multimedia applications based on the captured hand poses. 
    For example, \cite{zhang2022mid} proposed an end-to-end black-box mid-air gesture recognition method for in-vehicle media interaction and control with eye movement based experiments. 
    \cite{sluyters2023consistent} designed a consistent mid-air gesture interaction on a large display for browsing multimedia objects, controlling the selected target and volume, and more. 
    Black-box gesture interaction methods are used by many interaction tasks due to their fast and direct feedback performance but are difficult to apply in tasks that require interactive interpretability or visualization requirements. 

    This study focuses on the more challenging topic of mid-air hand pose embedding and interaction.
    Challenges in adopting this potentially more intuitive and convenient approach to multimedia control are due to the complexity and challenges of hand-pose embedding. 
    The complexity is mainly due to the inherent flexibility of human hands, resulting in a wide variety of complex postures. The flexibility and freedom of hand postures are fundamentally crucial considerations in designing interactions. 
    Unlike virtual reconstruction of hand poses directly using external handheld devices or radar \cite{bhardwaj2021tangibledata}, encoding hand poses and visualizing their positions in low-dimensional space can be challenging \cite{rusu2022low} due to their high complexity.
    For instance, \cite{xu2023continuous} proposed to construct a sequential gesture-based embedding space for interaction with hand movement. 
    However, they mainly focus on the pipeline of interaction process, while ignoring the stability and smoothness of system interaction, and only considering fixed hand postures limits system flexibility.
    Because interacting in a visualized embedding space using hand-pose movement has many uncontrolled factors, such as unavoidable physiological jitter \cite{lobov2016human}, etc.

     In this work, we present a new multimedia adapter that acts as a virtual sensor, namely a hand-pose embedding interactive system, to provide a novel and instructive method for user-led multimedia interaction by hand pose movement.  It mainly contains an innovative augmented VAE-based system framework with multiple post-processing designs for stability and smoothness consideration. 
     Furthermore, to improve the user experience, we designed a user guidance window for user interaction based on hand pose reconstruction.
     Our contributions are as follows: 1) We introduce a novel virtual sensor \textbf{HpEIS}, a learnable hand-pose embedding interactive system, to provide new inspiration for interacting with multiple multimedia collections with flexible hand movement. 
     2) We proposed a novel augmented VAE model to encode the hand pose into a visualized latent space for interaction, which contains an innovative augmentation strategy with an anti-jitter regularisation term based on a pilot experiment. 
     3) We introduced stabilization post-processing and smoothing post-processing to specifically deal with instability due to physiological jitter and system sensitivity.
     4) We are the first to design a real-time user guidance window based on the hand pose reconstruction in the hand pose interaction system. This greatly improves the user interaction experience.
     5) Extensive user study experiments demonstrate substantial advancements in the user experience of our system in terms of stability and smoothness, while maintaining interaction flexibility. Furthermore, we demonstrate the feasibility and usability of connecting to multimedia via HpEIS.

\begin{figure}[t] 
	\centering
	\includegraphics[width=0.7\linewidth]{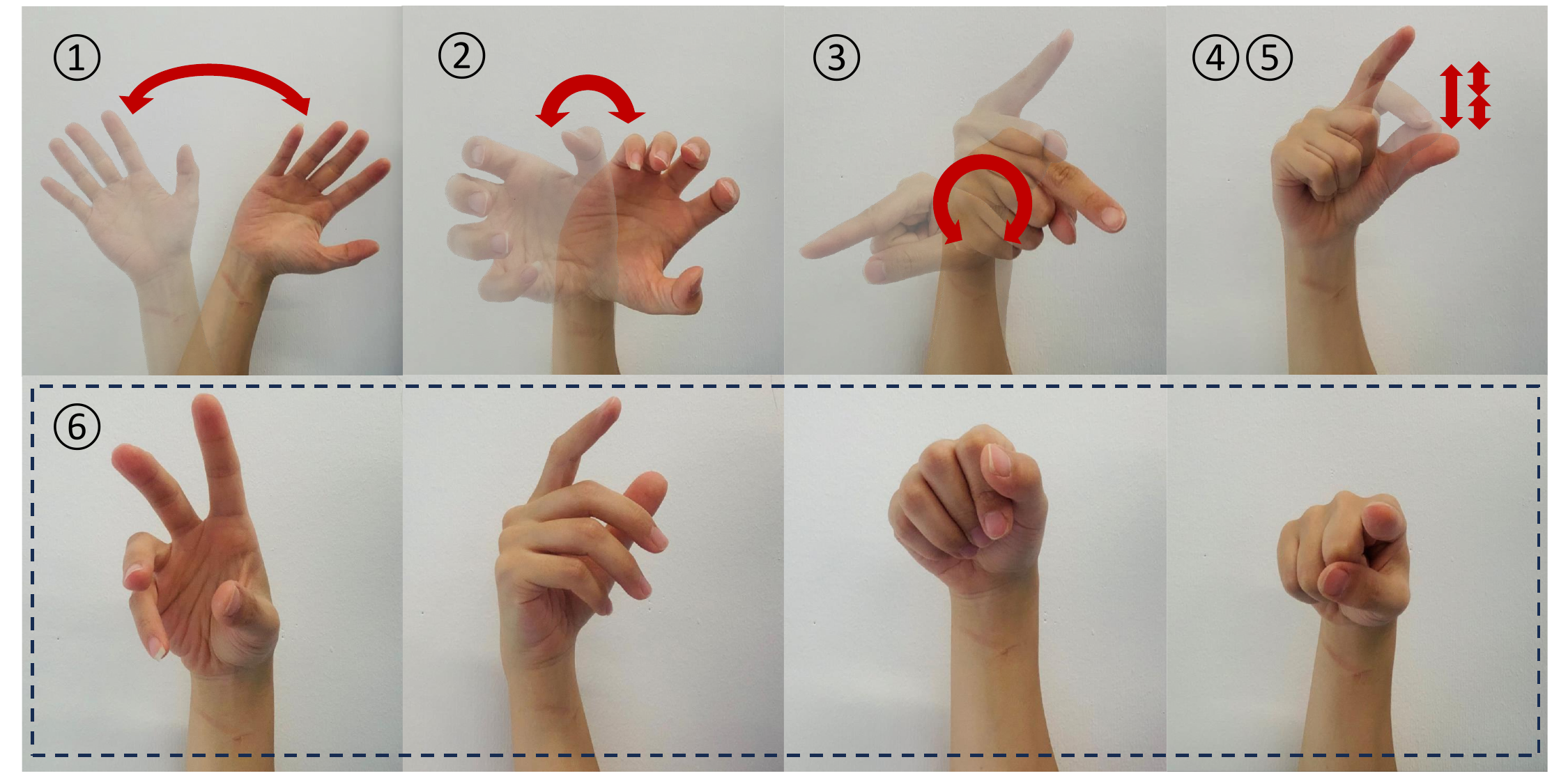}
	\vspace{-1em}
	\small\small\caption{Five functional hand-pose schematics and some generic movements. } 
	\label{fig:hand_data}
	\vspace{-1.0em}
\end{figure} 

\begin{table}[t!]
\centering
\scriptsize
\fontsize{8}{10}\selectfont
\renewcommand\tabcolsep{8.0pt}
\vspace{-0.5em}
\caption{Overview of hand-pose frames corresponding to Fig. \ref{fig:hand_data}.} 
\label{tab:gesture}
\vspace{-1em}
\resizebox{\columnwidth}{!}{%
\begin{tabular}{c|ccccc}
\hline
No.   & \multicolumn{1}{c|}{Hand Pose}  & \multicolumn{1}{c|}{Total} & \multicolumn{1}{c|}{No.} & \multicolumn{1}{c|}{Hand Pose} & Total \\ \hline
\ding{172}   & \multicolumn{1}{c|}{palm sweep} & \multicolumn{1}{c|}{21803} & \multicolumn{1}{c|}{\ding{175}} & \multicolumn{1}{c|}{pinch}     & 9181  \\
\ding{173} & \multicolumn{1}{c|}{hand dial rotation}  & \multicolumn{1}{c|}{18912} & \multicolumn{1}{c|}{\ding{176}} & \multicolumn{1}{c|}{double pinch}     & 8980 \\
\ding{174} & \multicolumn{1}{c|}{index finger circle} & \multicolumn{1}{c|}{20974} & \multicolumn{1}{c|}{\ding{177}} & \multicolumn{1}{c|}{generic movement} & 9781 \\ \hline
Total & \multicolumn{5}{c}{89631}                                                                                                        \\ \hline
\end{tabular}%
}
\vspace{-2.2em}
\end{table}

\section{Data Collection}

 10 participants (6 male, 4 female) between the ages of 22 and 30 were invited to collect hand pose data. They were all right-handed users with normal hand movements. 
 The camera used to capture hands was the RGB camera of Intel\textsuperscript{\textregistered} RealSense$^\text{TM}$ LiDAR Camera L515 depth camera (frame rate is 30 fps). 
 We introduce pilot experiments with one expert user.
 12 participants (7 men and 5 women) participated in our User Study.
 Additionally, 12 participants were divided equally into two groups. The first group experimented without a user guidance window, and the other group was guided with one.

Fig. \ref{fig:hand_data} presents the collected hand poses for interaction, including five functional hand poses, called palm sweep, hand dial rotation,  index finger and thumb pinch, and index finger and thumb double pinch. 
In addition, we incorporate generic hand pose movements into the mid-air hand-pose embedding space to enhance the flexibility of the system interaction. 
These hand poses encompass multiple angular relationships between the thumb and index finger, as well as the relationship between different inter-articular angles and flexion changes of a single finger or two fingers. Tab. \ref{tab:gesture} counts the details of collections. In total, we collected 89,631 frames of hand pose.

\section{Overview of Interaction System}

\begin{figure}[t] 
    	\centering
    	\includegraphics[width=0.75\linewidth]{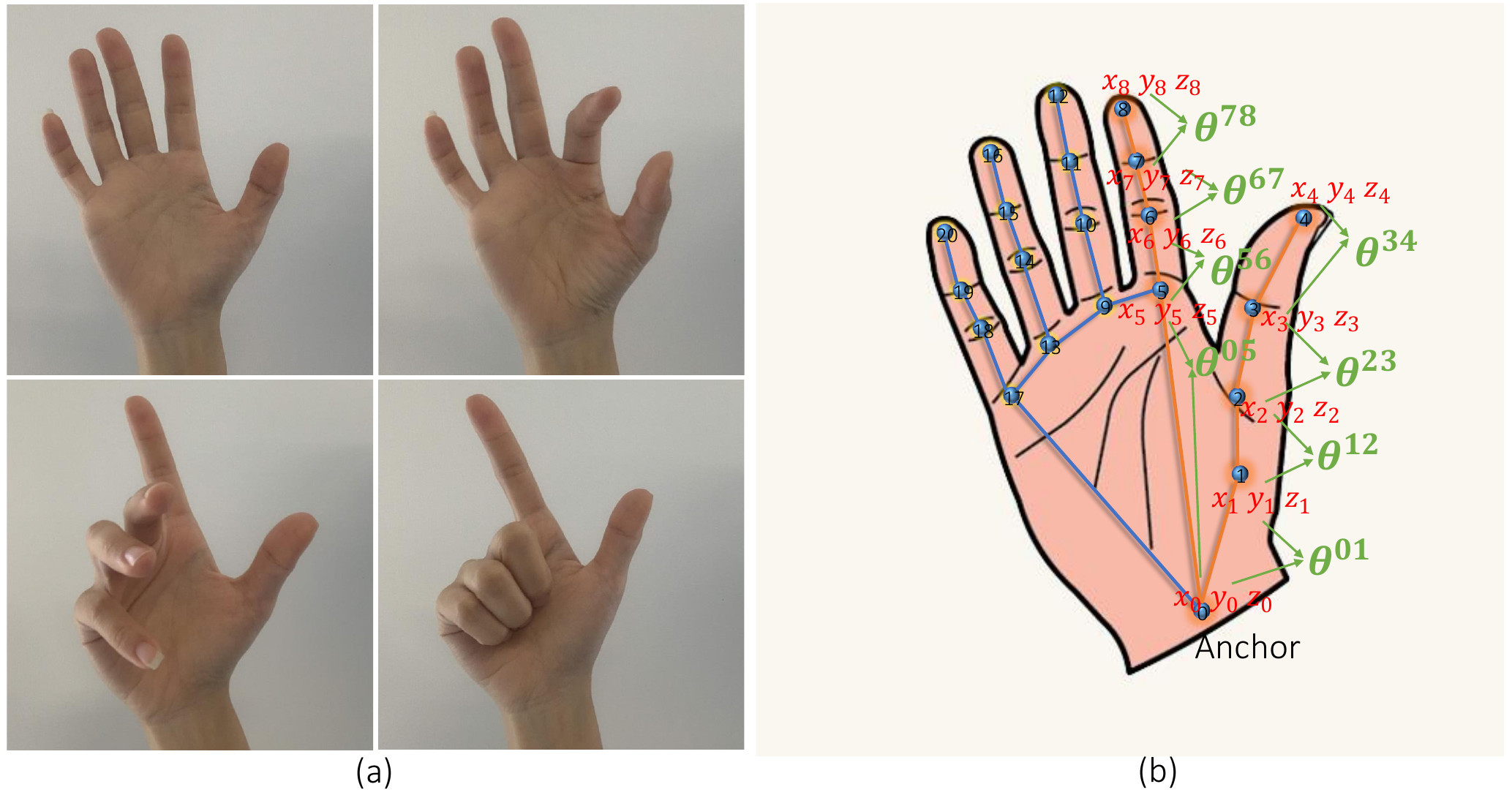}
    	\vspace{-1.2em}
    	\small\small\caption{Example of MediaPipe extraction and quaternion conversion. (a) means that we only consider the thumb and index fingers and their angles with the wrist, and the posture of other fingers does not affect the hand-pose representation. (b) means the quaternion conversion for 9 detected landmarks.} 
    	\label{fig:hand}
    	\vspace{-1.6em}
    \end{figure} 
\subsection{Preliminaries}
    In this work, as shown in Fig. \ref{fig:hand} (a), we use the thumb and index finger to form a hand pose for interaction and consider the relative angle information of both fingers to the wrist simultaneously.  
    Note that, different poses of the remaining fingers do not affect the mid-air hand-pose definition, thus maintaining a certain degree of hand flexibility and pose robustness.
    To get the hand-pose embedding space, a camera is used to capture the hand inputs $\mathcal{G}=\{g_1, \dots, g_n\}$, where $n$ is the length of the hand movement sequence of each user, and employ MediaPipe \cite{lugaresi2019mediapipe} to detect the 21 key points $\mathcal{P} = \{p_1,\dots,p_{21}\}$ for each hand frame, where each point has three-dimensional coordinates $ p_i = (\hat{x}_i, \hat{y}_i, \hat{z}_i)$. 
    We find that different experimental users with different hand sizes at different distances from the camera device usually produced different interaction results using mid-air hand poses with identical meanings in the embedding space. 
    To address this issue, we employ quaternion \cite{rieger2004systematic} to transfer the hand key landmarks, avoiding the effect of hand size, hand position in the sensor field and distance from the sensor.
    Specifically, as shown in Fig. \ref{fig:hand} (b), we choose an anchor point at the wrist and 8 landmarks at the joints of two fingers, to calculate the quaternion number of adjacent landmarks. 
    After transferring, we obtain new quaternion-based rotation angle hand-pose movement representations $\mathcal{Q} = \{q_1, \dots, q_n\}$ based on the detected landmarks, where $q_i=[\theta_i^{01},\theta_i^{12},\theta_i^{23},\theta_i^{34},\theta_i^{05},\theta_i^{56},\theta_i^{67},\theta_i^{78}]$ for each hand frame. 
    Afterwards, we use Variational Autoencoder (VAE) \cite{kingma2014auto} to encode and decode all hand pose representations and use the intermediate 2-dimensional latent representations as coordinates to compose the embedding space $\mathcal{S}=\{s_1, \dots, s_m\}$ for interaction and visualization, where $s_i=(x_i, y_i)$. 
    Finally, we can obtain a 2-dimensional hand-pose embedding space $\mathcal{S}$, which allows users to explore in all directions with different hand postures. 
    Additionally, we provide new users with visualized guidelines based on the decoder in trained VAE for hand pose movement, while also reducing user discomfort once the embedding space is updated.

\subsection{Mid-air Hand-pose Embedding Space } \label{space}
\begin{figure}[t] 
	\centering
	\vspace{-0.3em}
	\includegraphics[width=1\linewidth]{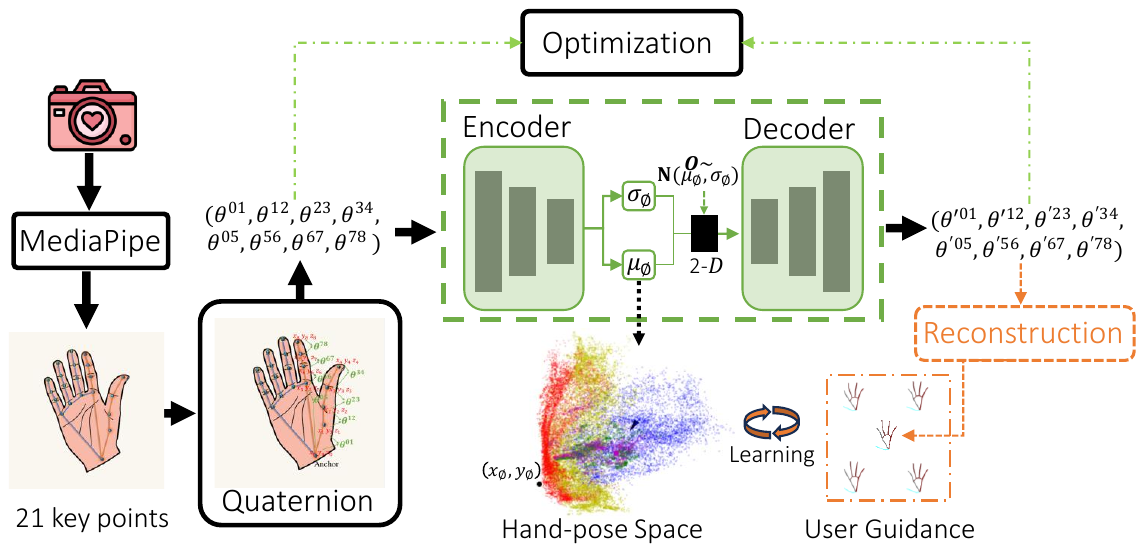}
	\vspace{-2.5em}
	\caption{The pipeline of our mid-air hand-pose embedding space construction and user guidance reconstruction.} 
	\label{fig:vae}
	\vspace{-1.6em}
\end{figure}

    Fig. \ref{fig:vae} shows our mid-air hand-pose embedding space construction pipeline.
    For each hand frame $g_i$, we use a VAE \cite{kingma2014auto} to obtain the latent 2-D embedding as the pose coordinate $s_i=(x_i, y_i)$, which conforms to a learnable Gaussian distribution with mean $\mu$ and standard deviation $\sigma$ with quaternion-based rotation angle representation $q_i\in \mathrm{R}^8$ input. 
    It contains a Multi-Layer Perceptron (MLP) based encoder and an MLP-based decoder.  
    Specifically, each MLP contains 4 fully connected layers, where the neuron numbers in encoder are 128,
    96, 64 and 2, respectively and the reverse in the decoder. The 2-D latent embeddings are normalized between 0 and 1.
    
    Our goal is to make the sample representations before encoding and after decoding as similar as possible by optimization, and the latent space used for decoding obeys the Gaussian distribution.
    In other words, the Mean Squared Error (MSE) between input $q_i=[\theta_i^{01},\theta_i^{12},\theta_i^{23},\theta_i^{34},\theta_i^{05},\theta_i^{56},\theta_i^{67},\theta_i^{78}]$ and output $q'_i=[\theta_i^{'01},\theta_i^{'12},\theta_i^{'23},\theta_i^{'34},\theta_i^{'05}, \theta_i^{'56},\theta_i^{'67},\theta_i^{'78}]$  and the  Kullback Leibler(KL) divergence of the distribution of low-dimensional data and the standard normal distribution are as small as possible, which are as follows:
    \begin{equation}
    \small
    \begin{aligned}
        \mathcal{L}_{MSE} = \frac{1}{8} \sum_{j=1}^{8}{\left(\theta_i^{(j)}-\theta_i^{'(j)}\right)}^2,
    \end{aligned}
    \end{equation} 
    \begin{equation}
    \small
    \begin{aligned}
        \mathcal{L}_{KL}\left( P_{\phi}(O|q_i)||P{\varphi}(O)\right) = \mathcal{L}_{KL}{\left(\mathcal{N}(\mu_\phi,\sigma_\phi),\mathcal{N}(0,1)\right)} \\
       =  - \frac{1}{2}\sum_{j=1}^{2}\left( 1+ \log{\left(\left(\sigma_{\phi}^{(j)}\right)^{2}\right)} 
          - \left(\sigma_{\phi}^{(j)}\right)^{2}-\left(\mu_{\phi}^{(j)}\right)^{2}\right), 
     \end{aligned}
    \end{equation} 
    where $P_{\phi}$ is a binary independent Gaussian distribution with learnable mean $\mu_i$ and standard deviation $\sigma_i$, $P_{\varphi}$ is a binary standard Gaussian distribution $\mathcal{N}(0,1)$. $O$ is a randomly re-sampled 2-D latent feature from $P_\phi$ to decode. Finally, we use binary mean $\mu_\phi$ as the coordinate $(x_i, y_i)$ to present the hand-pose position in the embedding space. 

\subsection{User Guidance Window} \label{guidance}
    To bring users, especially novice users, a novel and learnable exploration and interaction experience, we provide an innovative method called user guidance window in Fig. \ref{fig:vae}, which uses a trained decoder to obtain hand-pose decoding representations in four directions around the current hand pose and inversely uses quaternion to reconstruct the image expression of these neighbour hand poses. 
    Specifically, given the current mid-air hand pose $g_i$, we use the trained encoder to obtain the means $(x_\mu, y_{\mu})$ of latent representation as the current position in the hand-pose space.
    Then, we sample 20 latent representations with the same distribution as the current hand pose as neighbour hand poses, and take 4 latent representations in four directions as the final hand-pose guidance inputs to the trained decoder for decoding and reconstruction. 
    During the reconstruction process, we invert the quaternion and recalculate the positional relationship of the thumb, index finger and wrist, according to the decoded quaternion-based neighbour hand-pose representations $\{q'_1,q'_2,q'_3,q'_4\}$. 
    In addition, we selected eight orientations at the edge of the entire space for hand posture reconstruction as fixed orientation guidance. 
  
\begin{figure}[t] 
	\centering
	\vspace{-0.3em}
	\includegraphics[width=1\linewidth]{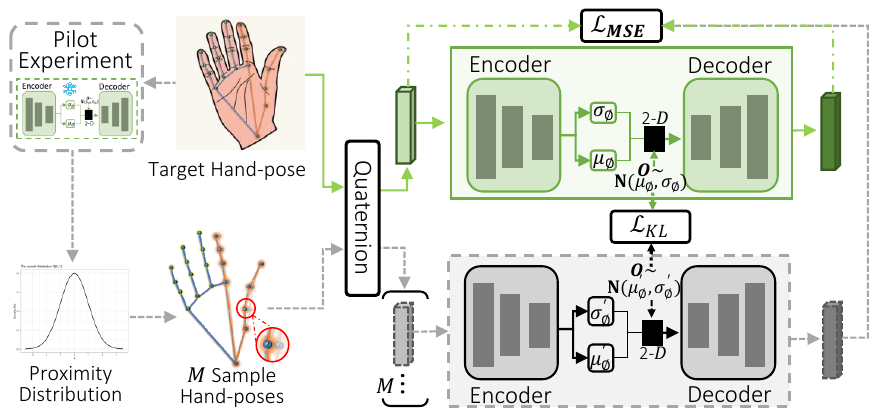}
	\vspace{-2em}
	\caption{The data augmentation strategy and the VAE re-training process with stability constraints.} 
	\label{fig:augmentation}
	\vspace{-2em}
\end{figure}   

\subsection{Design for System Stability and Smoothness} \label{stab}
     We conducted pilot experiments (in Section \ref{pilot}) on the current interactive system. However, we found two intuitive instability issues during our pilot experiments, which are (1) hand-pose embedding space is overly sensitive to subtle hand changes caused by physiological jitter and (2) space track exhibits jump due to error detection of hand key landmarks from intrinsic defects of the hand detector (MediaPipe). 
     To improve the interactive stability of the VAE-based hand-pose embedding space, we then design an augmentation strategy for VAE model with stability and smoothness post-processes.

    \textbf{Augmentated VAE Model.}
        Inspired by \cite{sinha2021consistency}, we design a new data augmentation strategy in Fig. \ref{fig:augmentation} based on pilot experiments to mitigate the position oscillations in embedding space due to hand jitter and add a new optimization objective function for this purpose to refine and retrain the VAE. 
        Specifically, we first trained a pilot model on collected hand poses. Then we designed a pilot experiment (in Section \ref{pilot}) to collect the real-interaction data, which contains 10 poses closest to each target pose location as the stable proximity hand poses. 
        After that, we can get the proximity distribution for each landmark of the target hand pose based on the collected proximity hand poses. We sample M (M=100) new proximity hand poses as the augmented hand poses, where the mean is from the target feature and the standard deviation is averaged from proximity hand poses. 
        Finally, we use quaternion to convert the target and augmented hand-poses landmarks and input them into the new VAE. 
        Our goal is to overcome the system sensitivity caused by slight movement or jitter by optimizing the minimum distance between the jittered hand poses and the target hand pose and the space distribution of both. The objective functions between the target hand-pose representation $q_i$ and each augmented representation $q_m$ can be described as:
        \begin{equation}
        \small
        \begin{aligned}
                \mathcal{L}_{MSE} = \frac{1}{8} \sum_{j=1}^{8}{\left(\theta_i^{(j)}-\theta_m^{(j)}\right)}^2 ,
        \end{aligned}
         \end{equation}
        \begin{equation}
        \small
        \begin{aligned}
            & \mathcal{L}_{KL}{\left( P_{\phi '}(O'|q_m)||P_{\phi}(O|q_i)\right)} = \\ & - \frac{1}{2}\sum_{j=1}^{2}{\left( 1 - \log{{\left( \frac{\sigma_{\phi}^{(j)}}{\sigma_{\phi '}^{(j)}} \right)^{2} }} 
              - \frac{\left(\sigma_{\phi '}^{(j)}\right)^{2} + \left(\mu_{\phi'}^{(j)}-\mu_{\phi}^{(j)}\right)^{2}}{ \left( \sigma_{\phi}^{(j)} \right) ^{2}} \right)}. 
        \end{aligned}
        \end{equation}
        Where the $\mathcal{L}_{KL}$ is similar to \cite{sinha2021consistency} for minimizing the difference in distribution between training hand poses and proximity hand poses. Additionally, the new $\mathcal{L}_{MSE}$ between the proximity hand poses and the target hand pose serves as a regularization term to further limit the difference between the jittered hand pose and the target hand pose in reality. Notably, we finally minimize the objective functions containing the loss computation of VAE itself.

    \textbf{Stability Post-processing Design.}\label{stab_post}
        In order to solve the error detection issue caused by MediaPipe, for example, inaccurate landmark detection occurs when the finger is perpendicular to the camera (last image in the bottom row of Fig. \ref{fig:hand_data}) or when the finger is obscured (second image in the bottom row of Fig. \ref{fig:hand_data}), we design a post-processing measure to further improve the stability of system interaction. 
        Here we consider the Euclidean distance change of the coordinates in the embedding space to be greater than 0.1 as the mutation coordinates, i.e. unstable phenomenon, and there may exist error detection. 
        We neutralize the unstable point $s_{i}$ with the following operation:
        \begin{gather}
        \small
           s_{i}=\begin{cases}
               \frac{1}{2}(s_{i-1}+s_{i}) & , \mathrm{if} \sqrt{(s_{i-1}-s_{i})^{2} } >0.1 \\
              s_{i}                 & , \mathrm{if} \sqrt{(s_{i-1}-s_{i})^{2} } <0.1.  
            \end{cases}
        \end{gather}
    
\textbf{Smoothness Post-processing Design.} \label{smooth}
    To further improve the unsmooth experience caused by jitter near the target point and path mutation during interaction, we use One Euro Filters with different parameters, i.e. cutoff frequency $CF$ and slope threshold $ST$, to handle different hand-pose interaction scenarios separately. 
    Specifically, we divide user interaction scenarios into two types, fast-moving interaction and slow-moving interaction. The former tends to occur during the initial stage of the interaction, where the user's hand moves faster, so the hand moves a greater distance between frames. Conversely, the latter tends to occur at the end of the interaction near the target point, where the user's hand moves slowly and the movement distance is small.
    Furthermore, we argue that users value the speed of exploring to the target point at the initial stage of interaction, but value the accuracy of reaching the target point at the end of the interaction stage. 
    To this end, according to the distance between two neighbouring hand-pose points in embedding space, we choose two different group parameters for our filter, as follows:
    \begin{gather}
    \small
       (CF, ST) =\begin{cases}
           (0.04, 0.85)   & ,\mathrm{if}\ D(g_{i}- g_{i-1}) >0.015\\ 
           (0.005,0.75)   & ,\mathrm{if}\ D(g_{i}- g_{i-1}) <0.015. 
        \end{cases}
    \end{gather}
    Note that, due to the consideration that real-world hand movements are less sensitive than corresponding movements in embedding space, we use the movement distance between two neighbouring hand poses, $g_{i-1}$ and $g_{i}$, calculated from 9 landmarks of two fingers in the real world as the parameter setting condition.

\section{Experimental Study}

\subsection{Pilot Experiments} \label{pilot}
    We conduct pilot experiments to identify problems, refine research designs and assess feasibility. 
    The hand movement plane was specified to be within 30-60 centimetres from camera. The hand starting position and starting pose were not fixed. Finally, when the user finds the target point, a Bluetooth knob (Griffin PowerMate) was used as a labelling tool and then headphones produce audio feedback.
    
       \textbf{Baseline Experiment.}
       According to Section \ref{space}, we trained our VAE on the collected hand poses as the pilot model. 
        Then an expert user conducted an exploration of the hand-pose embedding space.  
        Our goal was to verify whether the user could find the target point in the 2D embedding space and to detect the interaction problems during process. 
        Firstly, we randomly select 10 target points with different colours for the pilot user in the trained embedding space.  Different coloured points encoded in the embedding space mean the collected hand-pose embedding points. 
        The pilot user was presented with a black circular pointer representing the current hand-pose position in the space. 
        Then, we asked the pilot user to press and hold the knob when he intuitively thought he had found a point to mark the point found, and the system would display the next target point. 
        We evaluate the interactive usability of our system by the distance (between the final point and the target point) and the time taken (between the start and the stop) when the user thought he had reached the target point within a limited time.

        \begin{table}[]
        \scriptsize
        \begin{center}
        \fontsize{8}{10}\selectfont
        \renewcommand\tabcolsep{7.2pt}
        \caption{Uncertainty comparison of models. }
        \label{tab:results}
        \vspace{-2em}
        \resizebox{\columnwidth}{!}{%
        \begin{tabular}{c|c|c|c|c}
        \hline
        Model Name   & \begin{tabular}[c]{@{}c@{}}Stable \\ post process\end{tabular} & \begin{tabular}[c]{@{}c@{}}Smooth \\ post process\end{tabular} & UPA     & USA     \\ \hline
        Basic VAE    &                           &                           & 0.01925          & 0.00317          \\
        StdAugVAE    &                           &                           & 0.11750          & 0.01560          \\
        AvgStdAugVAE &                           &                           & 0.00910          & 0.00225          \\
        AvgStdAugVAE & \checkmark                &                           & 0.00343          & \textbf{0.00084} \\
        AvgStdAugVAE &                           & \checkmark                & 0.00409          & 0.00092          \\
        AvgStdAugVAE & \checkmark                & \checkmark                & \textbf{0.00223} & 0.00101          \\ \hline
        \end{tabular}%
        }
        \end{center}
        \vspace{-3em}
        \end{table}

\begin{figure}[t] 
        	\centering
        	\includegraphics[width=0.9\linewidth]{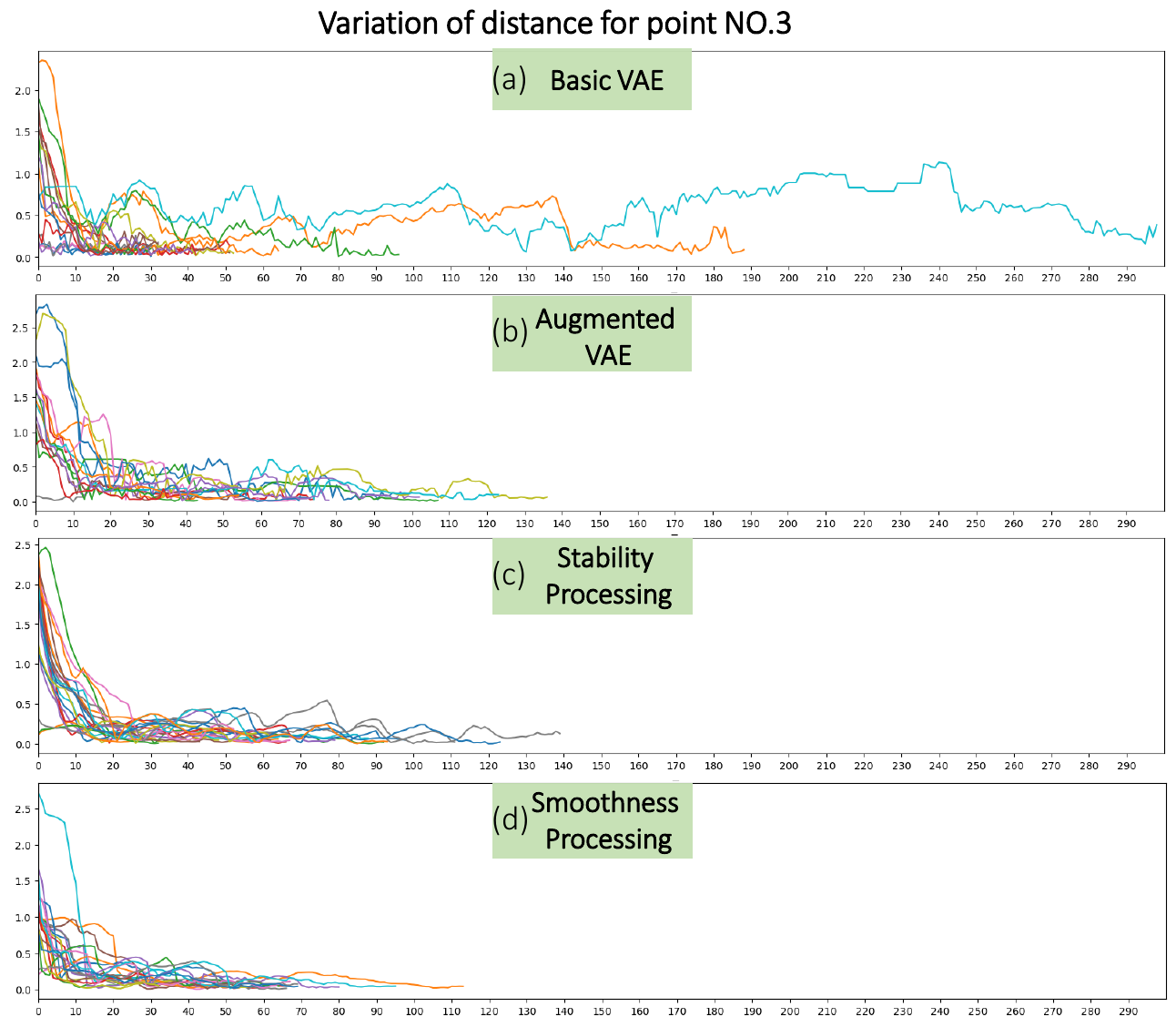}
        	\vspace{-1.3em}
        	\caption{Distance changes for each attempt of one user in the pilot experiments in relation to time on one random target point. The horizontal axis represents time (s/10) and the vertical axis represents Euclidean distance.} 
        	\label{fig:post_process}
        	\vspace{-1.8em}
        \end{figure}

       
        After the pilot experiments, we computed the uncertainty of the last ten frames of each attempt around different target points. 
        Tab. \ref{tab:results} shows the average uncertainty in principal axes (UPA) and the average uncertainty in secondary axes (USA) for the last ten frames of the 2D representation of hand movements near the target point. 
        Fig. \ref{fig:post_process} (a) showed the distance of each hand pose from the target points and the time-consuming. 
        Two obvious issues can be found: (i) it is harder to anchor the hand pose to the target point, i.e. the pose uncertainty in the target area is high in the first row of Tab. \ref{tab:results}; (ii) the instability and non-smoothness of the user's interaction, i.e. the oscillation of the distance from the target points in Fig. \ref{fig:post_process} (a).
        We believe that these two problems are mainly caused by physiological jitter and system sensitivity, and the statistical plots can be found in the Appendix.

        \textbf{Augmentation Experiment.}
        To address the instability and sensitivity, we explored two augmentation strategies in Section \ref{stab}. (i) \textbf{StdAug VAE}: For the last ten frames of 200 explorations (target points are not fixed), we calculate the standard deviation of each coordinate of each point on their original landmarks, and finally average the standard deviation of each target point and then sample.
        (ii) \textbf{AvgStdAug VAE}: Different from StdAug VAE, we first calculate the standard deviations of the last 10 frames during each exploration and then calculate the average standard deviations of the 200 explorations. 
        By comparing different models in Tab. \ref{tab:results},  StdAug VAE make the distribution of hand poses more uncertain but AvgStdAug VAE is better, i.e. the smallest uncertainty in both principal and secondary axis directions.
        We believe that the AvgStdAug VAE can separate different search processes and determine the computational uncertainty more accurately than the average uncertainty for all attempts in the StdAug VAE.
        Additionally, we also displayed the distance of each interaction from the target points and the time from the start points in Fig. \ref{fig:post_process} (b). 
        Compared with basic VAE (Fig. \ref{fig:post_process} (a)), our AvgStdAug VAE with anti-jitter KL loss can better improve the stability of interactions when approaching target points.

        \textbf{Stability Experiment.}
        To further improve the stability, we further introduced a stable post-processing design in Section \ref{stab_post}. 
        By comparing Fig. \ref{fig:post_process} (b) and Fig. \ref{fig:post_process} (c), we can find that the stable post-processing design can better improve the stability during interactions.
        In particular, Fig. \ref{fig:post_process} (b) shows many sudden increases or decreases of distance variation, i.e., some convex and concave cusps in (b), which are mostly higher than 0.1. After our stability post-process, a large proportion of the cusps have been replaced by more gentle curves in Fig. \ref{fig:post_process} (c). These changes are particularly evident in the distance variation curves after 6s for both types of users. 
        It reflects that our proposed stable post-processing has an excellent capability to improve the stability of system interaction.
        In terms of uncertainty in Tab. \ref{tab:results}, using stable post-processing reduces the results of UPA and USA by 62.7\% and 62.3\%, respectively, compared to the model without stable post-processing.
    \begin{table}[t]
        \scriptsize
        \begin{center}
        \fontsize{7}{9}\selectfont
        \renewcommand\tabcolsep{9.0pt}
        \caption{The results of different trade-offs of ($CF$, $ST$).}
        \label{tab: smooth results}
        \vspace{-2em}
        \resizebox{\columnwidth}{!}{%
        \begin{tabular}{c|c|ccc}
        \hline
        Model  & (CF, ST) set & UPA    & USA    & Time (s) \\ \hline
        Stable & -           & 0.00343 & 0.00084 & 4.9      \\ \hline
        Smooth &
          \begin{tabular}[c]{@{}c@{}}(0.040, 0.85)\\ (0.005, 0.75)\end{tabular} &
          0.00409 &
          \textbf{0.00092} &
          5.1 \\ \hline
        \begin{tabular}[c]{@{}c@{}}Stable + \\ Smooth\end{tabular} &
          \begin{tabular}[c]{@{}c@{}}(0.010, 0.80)\\ (0.004, 0.70)\end{tabular} &
          0.00461 &
          0.00094 &
          5.8 \\ \hline
        \begin{tabular}[c]{@{}c@{}}Stable +\\ Smooth\end{tabular} &
          \begin{tabular}[c]{@{}c@{}}(0.040, 0.85)\\ (0.005, 0.75)\end{tabular} &
          \textbf{0.00223} &
          0.00101 &
          \textbf{4.9} \\ \hline
        \end{tabular}%
        }
        \end{center}
        \vspace{-2em}
        \end{table}

        \textbf{Smoothness Experiment.}
        We proposed the smoothness experiment to address the phenomenon of non-smoothness according to Section \ref{smooth}. 
        By implementing smooth post-processing, the distance-time curves between the gesture-embedding position and the corresponding target point in the space have been significantly improved, as shown in Fig. \ref{fig:post_process} (d). 
        Due to the staged filtering effect of the One Euro Filter with different parameters, subtle wave peaks during the interaction process were significantly reduced, resulting in a smoother trajectory and stability near the target point. 
        Moreover, the different filter settings in the first and second stages of interactions allow for a better balance between speed considerations in the first stages of interactions and accurate considerations at the end of interaction. 
        We give the results of UPA, USA, and averaging time under different (CF, ST) trade-offs as shown in Tab. \ref{tab: smooth results}.
        In Tab. \ref{tab:results}, although the best results are achieved by USA for only adding stable post-processing, considering UPA and USA together, we believe that the addition of the two post-processes has a significant effect on the reduction of uncertainty.

    \begin{figure}[t] 
    	\centering
    	\includegraphics[width=0.8\linewidth]{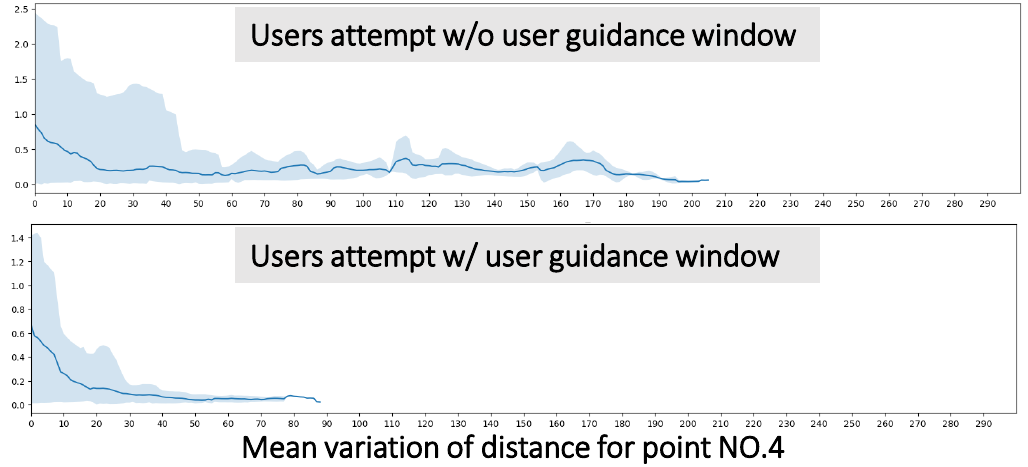}
    	\vspace{-1em}
        \caption{The average time-distance curves for the two groups of participants. The horizontal and vertical axes are the same as in Fig. \ref{fig:post_process}.} 
    	\label{fig:user_study_1}
    	\vspace{-2em}
    \end{figure} 

    \begin{figure}[t] 
    	\centering
    	\includegraphics[width=0.9\linewidth]{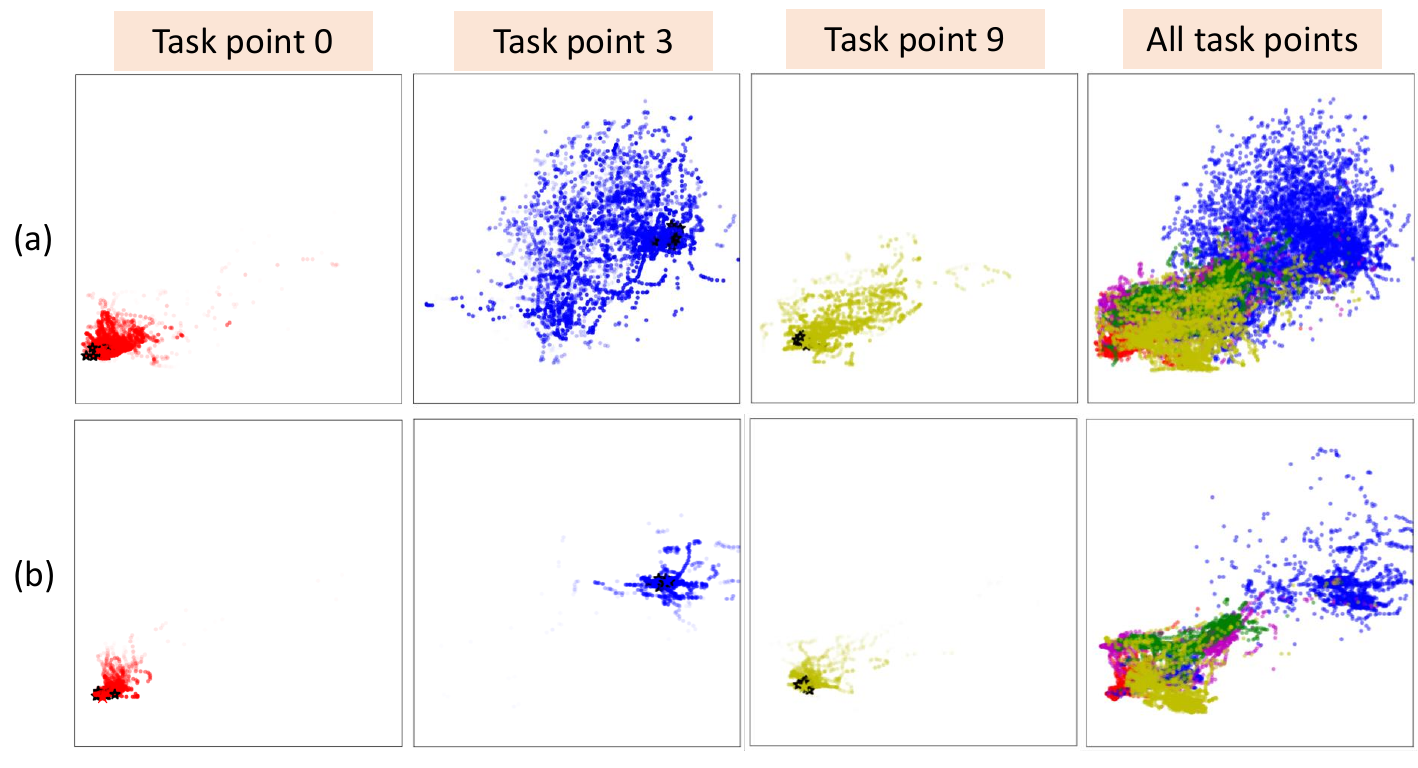}
    	\vspace{-1.0em}
    	\caption{Two random participants' hand movement trajectory points for some sample target points were visualized, (b) used the user guidance window and (a) did not. The black stars are stopping points of the final hand poses from a random participant. The colour from light to dark indicates the beginning to end of each attempt.} 
    	\label{fig:user_study_points}
    	\vspace{-1em}
    \end{figure}    
\subsection{User Study: System and Interface Test}

    \textbf{Experiment Design.}
    We conducted a user study to evaluate the performance of the proposed HpEIS. 
    Two versions of HpEIS are used, with the main difference being whether to use the innovative user guidance window based on hand pose reconstruction by the VAE decoder (in Fig. \ref{fig:vae}). 
    The task is the same as our pilot experiment, i.e. finding specific target points. 
    12 participants were evenly split into two non-overlapping groups to mitigate the influence of experiential factors. The first group completed the experiment without a user guidance window, while the second group utilized a user guidance window. 
    Next, participants started a user test in which each user conducted a total of 100 interactive explorations with a random target point from 10 specific points. 
    At the end of the experiment, participants filled out a questionnaire, and the corresponding statistical results are given in the Appendix.

    \textbf{Evaluation.}
    Similarly, the developed HpEIS is evaluated from two aspects, the duration of interactive exploration and the final distance from the target point.  
    Fig. \ref{fig:user_study_1} shows the average time-distance curves for the two groups of participants. In addition, we believe that the computationally averaged area (i.e., the light blue background area can somewhat indicate the stability and smoothness of the interaction.
    By comparing the results of the two groups of experiments, it is clearly observed that using the user guidance window can significantly reduce the user's approaching time to the target point, and the overall interaction process is more stable and smoother.
    Moreover, in Fig. \ref{fig:user_study_points}, we further visualized the location of the movement point of the hand pose for two random participants, where (b) used the user guidance window and (a) did not. 
    By comparison, the participant with user window guidance is easier to explore and faster to find a target. 
    This again demonstrates the effectiveness of the user guidance window.

     \begin{figure}[t] 
    	\centering
    	\includegraphics[width=1\linewidth]{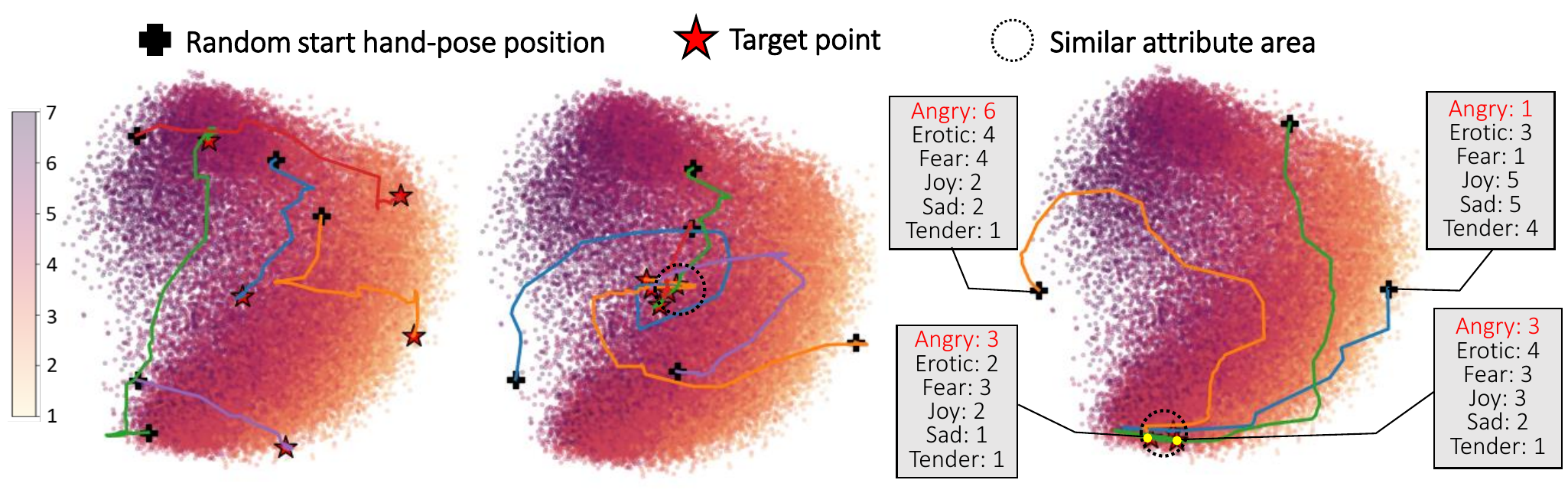}
    	\vspace{-1.5em}
    	\caption{A interactive application of HpEIS on a music multimedia space of expressive angry music (The darker color, the more intense the ``Anger" expressed of a song.). The first one is the exploration and interaction of music with different levels of anger. The latter two respectively represent multiple explorations and interactions corresponding to music with a similar attribute. Specifically, two yellow points are on a circle of similar attributes with a diameter of 0.1 in the music space, and the real distance in the high dimension is 1.00 and 2.24 respectively. The distance between the left start point and end points in higher dimensions is about 3.87, and the distance between the right start point and end points in higher dimensions is about 5.57, but only 0.47 and 0.65 in the music space. All distances are Euclidean distances.}  
    	\label{fig:music}
    	\vspace{-1.5em}
    \end{figure}        

\subsection{Application}
We believe the proposed HpEIS is a flexible multimedia adapter, which provides new inspiration for research on connecting multimedia.
In this study, we apply HpEIS to the music embedding space as an application example.
In particular, given a music-embedding space embedded from a similar trained VAE model, we normalize its extent to be the same as the hand-pose embedding space extent. 
After that, we can directly interact with the music-embedding space via hand movement. 
As shown in Fig. \ref{fig:music}, any hand pose can intervene in the space for interaction, thus ensuring interaction flexibility. 
For exploring music areas with similar attributes, we can use different starting hand poses to explore the target areas from different directions and paths, thereby maintaining the user's sense of novelty in the music system. 
Notable, the multimedia space is not limited to music space, any other multimedia space interaction can use the same HpEIS system. This improves the transferability of HpEIS and reduces repetitive work, which is not available in other interaction studies \cite{xu2023continuous}.

\section{Conclusion}
In this paper, we developed a novel HpEIS to adapt multimedia collections via mid-air hand movement using only a camera with MediaPipe software. HpEIS engages in (i) providing learnable and interpretable visible space exploration experiences, and (ii) providing flexible, stable, and smooth hand-pose interactions. 
To this end, we proposed the augmented VAE model encoding to obtain a 2D normalized hand-pose embedding space, which can be visualized for multimedia interaction via hand movement. 
A series of stability and smoothness post-processing operations have improved the overall user experience of the interactive system. 
We evaluated HpEIS in the task of finding the target points in the hand-pose embedding space with and without a novel user guidance window, as well as mapping music embedding space exploration. 
Comprehensive user experiments prove that our HpEIS not only provides a flexible, stable and smooth hand-pose interaction method, but also can inspire a new research direction in interacting with multimedia collections.

\bibliographystyle{IEEEtran}
\bibliography{icme2023template}

\end{document}